\pdfoutput=1

\documentclass[11pt]{article}

\usepackage{acl}

\usepackage{times}
\usepackage{latexsym}

\usepackage[T1]{fontenc}

\usepackage[utf8]{inputenc}

\usepackage{microtype}

\usepackage{amsmath}

\usepackage{graphicx}
\usepackage[noabbrev,capitalize]{cleveref}
\usepackage{caption}
\usepackage{subcaption}
\usepackage{algorithm}
\usepackage{algorithmic}
\usepackage{array, multirow, bigdelim, makecell, booktabs} 
\usepackage{amssymb}
\usepackage{adjustbox}
\usepackage{paralist}
\usepackage{lettrine}

\def\BH#1{{\color{red}BH: \it #1}}
\def\BH#1{}
\def\SS#1{{\color{blue}Sukanta: \it #1}}
\def\SS#1{}

\usepackage[normalem]{ulem} 
\def\XXX#1{\textcolor{red}{XXX #1}}
\def\XXX#1{}

\title{Simultaneous Translation for Unsegmented Input: \\ A Sliding Window Approach}

\author{Sukanta Sen\thanks{\hspace{4pt}Work performed while at the University of Edinburgh} \\
  University of Edinburgh \\
  sukantasen10@gmail.com \\\And
 Ond\v{r}ej Bojar \\
  Charles University \\
  MFF \'{U}FAL \\
  bojar@ufal.mff.cuni.cz \\\And
  Barry Haddow\\
  University of Edinburgh\\
  bhaddow@ed.ac.uk}

\begin{document}
\maketitle
\begin{abstract}
In the cascaded approach to spoken language translation (SLT), the ASR output is typically punctuated and segmented into
sentences before being passed to MT, since the latter is typically trained on written text.
However, erroneous segmentation,  due to poor sentence-final punctuation by the ASR system, leads to degradation in translation quality, especially in the simultaneous (online) setting where the input is continuously updated.
\XXX{In the previous sentence, Ondrej would highlight the non-existence of sentences in spontaneous speech rather than the continuous updates. Continuous updates are a problem, yes, but an orthogonal one to poor sentence-final punctuation. The previous sentence introduces two orthogonal problems at once, kind of mixing them up.} \SS{Also, in the inference time, we are using online text-flow which means we are not handling unstable input. I am not sure how do I rephrase the last sentence without a risk. For now I am keeping as it is for the submission purpose.}
To reduce the influence of automatic segmentation, we present a sliding window approach to translate raw ASR outputs (online or offline) without needing to rely on an automatic segmenter. We train translation models using parallel windows (instead of parallel sentences) extracted from the original training data. At test time, we translate at the window level and join the translated windows using a simple  approach to generate the final translation. 
Experiments on English-to-German and English-to-Czech show that our approach improves 1.3--2.0 BLEU points over the usual ASR-segmenter pipeline, and the fixed-length window considerably reduces flicker compared to a baseline retranslation-based online SLT system.
\end{abstract}

\section{Introduction}
\label{sec:intro}

For machine translation (MT) with textual input, it is usual to segment the text into sentences before 
translation, with the boundaries of sentences in most text types indicated by punctuation. For spoken language
translation (SLT), in contrast, the input is audio so there is no punctuation provided to assist 
segmentation.
Segmentation thus has to be guessed by the ASR system or a separate component.
Perhaps more importantly, for many speech genres the input cannot easily be segmented into
well-formed sentences as found in MT training data, giving a mismatch between training and test.

In order to address the segmentation problem in SLT, systems often include a segmentation
component in their pipeline, e.g. \citet{cho2017nmt}. In other words, a typical \emph{cascaded} SLT
system consists of  automatic speech recognition (ASR -- which outputs lowercased, unpunctuated text)
a punctuator/segmenter (which adds punctuation and so defines segments) and an MT system. The
segmenter can be a sequence-sequence model, and training data is easily synthesised from 
punctuated text. However adding segmentation as an extra step has the disadvantage of introducing
an extra component to be managed and deployed. Furthermore, errors in segmentation 
have been shown to contribute significantly to overall errors in SLT \citep{li2021sentence}, 
since neural MT is known to be susceptible to degradation from noisy input  \cite{khayrallah-koehn-2018-impact}. 

These issues with segmentation can be exacerbated in the \emph{online} or \emph{simultaneous} setting. 
This is an important use case for SLT where we want to produce the translations from live speech, 
as the speaker is talking. To minimise the latency of the translation, we would 
like to start translating before speaker has finished their sentence. Some online low-latency ASR
approaches will also revise their output after it has been produced, 
creating additional difficulties for the 
downstream components. In this scenario, the segmentation into sentences will be more uncertain and we 
are faced with the choice of waiting for the input to stabilise (so increasing latency) or 
translating early (potentially introducing more errors, or having to correct the output
when the ASR is extended and updated).

To address the segmentation issue in SLT, \citet{li2021sentence} has proposed to a data augmentation technique which simulates the bad segmentation in the training data. 
They concatenate
two adjacent source sentences (and also the corresponding targets) and then start and end of the concatenated sentences are truncated proportionally.  

We use a \emph{sliding window} approach to translate unsegmented input. In this 
approach, we translate the ASR output as a series of overlapping windows, using a merging algorithm to turn the translated windows into a single continuous (but still sometimes updated) stream. The process is illustrated in \cref{fig:example:match:nomatch}. 
To generate the training data, we convert 
the sentence-aligned training data into window-window pairs, and remove punctuation and casing from 
the source. We explain our algorithms in detail in \cref{sec:window}. 

For online SLT, we use a \emph{retranslation} approach \citep{niehues2016inter,arivazhagan2019retranslation}, where the MT system retranslates a recent portion of the input each time there is an update from ASR. This approach has the advantage 
that it can use standard  MT inference, including beam search, and does not require
a modified inference engine as in streaming approaches (e.g.\ \citet{ma-etal-2019-stacl}). Retranslation
may introduce flicker, i.e. potentially disruptive changes of displayed text, when outputs are updated. Flicker can be traded off with latency by masking the 
last $k$ words of the output \citep{arivazhagan2019retranslation}.\footnote{This paper
also introduced the idea of \emph{biased beam search}, where the translation of an extended
prefix is soft-constrained to stay close to the translation of the prefix. Biased beam search
significantly reduces flicker, but it requires that ASR output has a fixed segmentation, and
uses a modified MT inference engine.}
Our sliding window approach is easily combined with retranslation to create an online SLT
system which can operate on unsegmented ASR. Each time there is an update from ASR, we retranslate the 
last $n$ tokens and merge the latest translation into the output stream. Using the fixed size window has 
the advantage of reducing flicker, since we control how much of the output stream can change on each
retranslation.

Experiments on English$\rightarrow$Czech and English$\rightarrow$German show that our sliding window approach improves BLEU scores for both online
and offline SLT. For the online case, our approach improves the tradeoff between latency and flicker. 


\BH{Do we need to explain how we are different to \citep{li2021sentence}?}
\XXX{OB: Not sure. If they do not merge windows at inference, I think we do not need to differentiate ourselves explicitly, because big enough difference is kind of implied.}

\def\ourvarsubscript#1#2{#1\textsubscript{#2}}  
\def\ourvarsubscript#1#2{#1$_{#2}$}  
\begin{figure}[t]
 \centering 
 \includegraphics[scale=0.5]{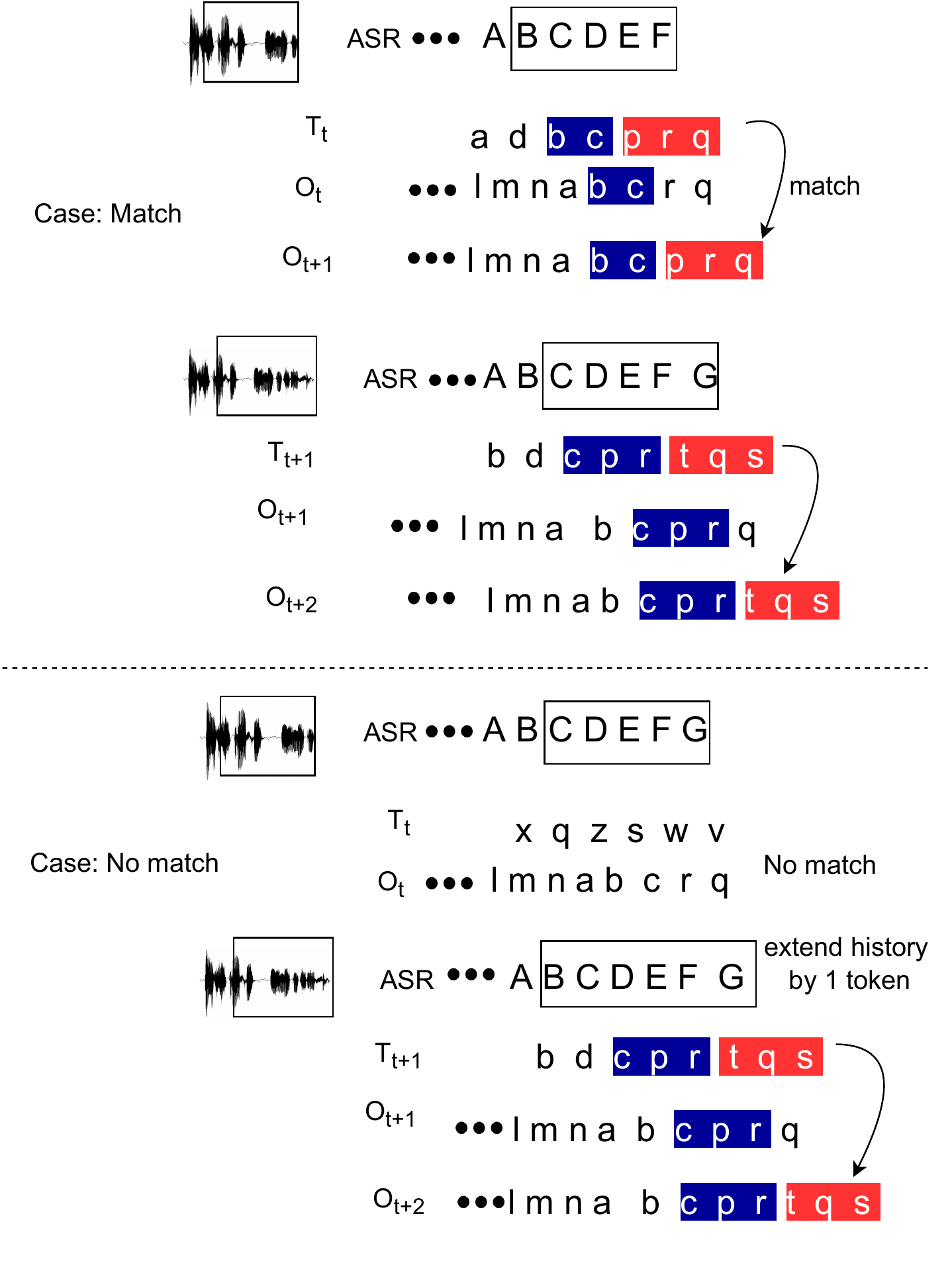}
 \caption{Example of how our proposed window-based translation works at test time in case of a match and no-match of translations of two subsequent windows. The text inside the rectangular box is the source window at time $t$, which is translated into output window
 (\ourvarsubscript{T}{t}) by the MT system. The text in blue (dark) shade shows the common segment between the output window (\ourvarsubscript{T}{t}) and the output stream (\ourvarsubscript{O}{t}) at time $t$. The text in red shade shows the segment newly added from the output window \ourvarsubscript{T}{t} into the output stream \ourvarsubscript{O}{t+1}. With no common segment between \ourvarsubscript{T}{t} and \ourvarsubscript{O}{t} (``No match''), we extend the input window into the history and translate again. $\bullet\bullet\bullet$ indicates there are more tokens. Note that we used characters here (instead of tokens) just for explanation.}
 \label{fig:example:match:nomatch}
\end{figure}

\section{Window-Based Translation}
\label{sec:window}
\subsection{Preprocessing}
\label{sec:preproc}

To make the parallel corpus resemble ASR output, we remove all punctuation (and other special
characters) from the source sentences and replace it with spaces. We then remove repeated spaces, and lowercase the source.



\subsection{Generating the Window Pairs for Training}
To convert the parallel corpus into a set of parallel windows, we use a word-alignment based approach. We first word-align the
pre-processed parallel corpus using \texttt{fast\_align} \citep{dyer-etal-2013-simple}, then we concatenate each side of the corpus to give two long 
lines. Note that the word alignments will however never cross sentence boundaries. We randomly select windows
of length 15--25 from the target side, and use the word alignment to get the corresponding source window. The algorithms are 
described in Appendix~\ref{app:window}.

A subtle detail is whether the original corpus was or was not shuffled at the level of sentences. An original, non-shuffled corpus provides the MT system with useful examples of cross-sentence conditioning, a very useful feature especially for spontaneous speech translation. A minority of our cs-en data is shuffled,
which adds some noise to the process, but our method works despite this. 
\BH{This raises the question of what happens if we remove the shuffled data. Something we could
address later}



\subsection{Translating Input Windows}
In our simultaneous MT setting, we assume that the ASR system is transcribing the incoming speech signal into a continuous stream of text.
To obtain a new input to the MT system, a fixed-side window is shifted by one token to right every time the stream is extended.
For every input window, the MT system translates it and sends it to the module that joins the output windows to the output stream as described in the next section.

\subsection{Joining the Output Windows}
\label{sec:joining}
Since
 two consecutive source windows overlap, the corresponding output windows normally have an overlap. We use this overlap to join an output window to the output stream.

We show the pseudo code for merging an output window to the output stream in Algorithm~\ref{algo:merge}. We assume that ASR produces an input stream $I$ which is continuously growing by one token at a time. Our algorithm requires a window length $w_l$, a threshold $r$ and the current output stream $O_t$. For every new token in $I$, our merge module in
Algorithm~\ref{algo:merge} is triggered. The MT system translates the last $w_l$ tokens of $I$ to a target window $T_t$. For any translated window $T_t$ and output stream $O_t$ at that time step, we find the longest common substring $s$. 
The threshold $r$ gives the required minimum length of the common substring. If the match is sufficiently long (``significant'' in the following), we merge the current target window $T_t$, otherwise we extend input window by 1 token to the left and translate again.

In our experiments, we extend the history to maximum of 5 tokens until we have found a significant match. A higher $r$ assures that the
translation of the current window will not accidentally match a random segment in the stream, and as the successive windows are just 1 token apart, we find a match almost always (see Appendix~\ref{app:match} for
details).  Once we have found a significant match, 
we merge $T_t$ with $O_t$ around the match, chopping the part of $T_t$ before match. 
This approach of joining windows is able to handle both the online and offline situations.




\def\abs#1{\:\lvert#1\rvert\:} 
\def\miniabs#1{\lvert#1\rvert} 
\begin{algorithm}[H]
\caption{Pseudo code for merging newly translated window into existing output.} 
\label{algo:merge}
\begin{algorithmic}[1]
\REQUIRE The current output stream $O_{t}$, input stream $I$, an $MT$ system, window length $w_l$, threshold $r \in (0, 1)$. 
\STATE $k = 0$ \COMMENT{{extra history considered}}
\WHILE{true}
\STATE $T_{t} \leftarrow MT(I[\abs{I}-(w_l+k) : \abs{I}])$ 
\STATE $O_t^{'} \leftarrow  O_{t}[\abs{O_t}- \abs{T_t} : \abs{O_t}] $  
    \STATE $s, i, j \leftarrow T_{t} \Psi O_{t}^{'}$ \COMMENT{$s$ is longest common substring. $i$ and $j$ are the start indices of match in $O_{t}^{'}$ and $T_t$} 
    \STATE $k \leftarrow k + 1$
    \IF {$\abs{s} \ge \abs{T_t} * r$ or $k > 5$}
     \STATE break
    \ENDIF
\ENDWHILE
\IF {$\abs{s} = 0$ } 
  \STATE {$i \leftarrow \abs{T_t}$}
  \STATE {$j \leftarrow 0$}
 \ENDIF
    \STATE $O_{t+1} \leftarrow O_{t}[0:\miniabs{O_t} - \miniabs{T_t} +i] + T_{t}[j:\miniabs{T_{t}}]$ 
\RETURN $O_{t+1}$

\end{algorithmic}
\end{algorithm}

\section{Datasets and Experimental Settings}
For training, we use parallel datasets from WMT 2020 \citep{barrault-etal-2020-findings} for English-German and from WMT 2021 \citep{akhbardeh-etal-2021-findings} for English-Czech (see Appendix~\ref{app:training} for details). 
For the validation set, we use the concatenation of IWSLT 2014,15 test sets for English-German, and newstest2019 
for English-Czech. We use the ESIC test set for evaluation. ESIC \citep{esic:2021} is a corpus derived from the European parliament proceedings which has transcripts of source English speech and interpreted German and Czech transcripts. This test set is aligned at document level.

We use the SentencePiece \cite{kudo2018sentencepiece} tokenizer for preprocessing the windows  with a shared subword \citep{sennrich-etal-2016-neural} vocabulary size of 32k. We train transformer-based\footnote{with 60 millions parameters. One model using 4 GPUs took on an average 2 days.} \citep{vaswani_attention_2017} NMT models using  the Marian  toolkit \citep{junczys-dowmunt-etal-2018-marian}. MT models are trained to convergence (using early stopping of 10) with a learning rate of $0.0003$, and translate using a beam of 6. We train the following two types of models: i) Baseline: trained on gold-segmented data and evaluated on segmented data generated by the ASR system; ii) Window: trained on windows of 10-25 tokens and evaluated on fixed length windows of ASR output.


\BH{We should explain where the ASR comes from}
\XXX{Ondrej thinks that the Niehues et al 2016 below is kind of sufficient reference.}
\SS{I think Barry is talking about the test set which we obtain after doing ASR and applying online text-flow (OTF). But I am not aware much about the ASR part. We are using OTF for getting the stable part.}


\section{Results}

\begin{table}[]
   \small
\setlength{\tabcolsep}{2.9pt}
    \centering
    \begin{tabular}{l|cc|ccccccc}
        \hline
        & \multicolumn{2}{c}{Baseline} & \multicolumn{7}{c}{Window} \\
        Pair & SF & SO & 8 & 10 & 12 & 14 & 16 & 18 & 20\\
        \hline
         en-de & 11.2 & 11.4 & 12.5 & 12.8 & 13.0 & 13.0 & 13.1 & 13.2 & 13.2\\ 
         en-cs & 9.4 & 9.4 & 10.0 & 10.3 & 10.4 & 10.5 & 10.6 & 10.6 & 10.7 \\ 
        \hline
    \end{tabular}
    \caption{Sacrebleu scores of segmented and window based approaches. SF: Offline segment level. SO: Online segment level.} 
    \label{tab:results:seg-vs-window}
\end{table}

We evaluate both the offline and online SLT. For offline SLT, the baseline system is trained using parallel sentences, and for the online version, the baseline system is a prefix-prefix retranslation system \citep{niehues2016inter,arivazhagan2019retranslation}. For our proposed window-based system, the offline and online are the same system.
We evaluate our proposed approach on ESIC using Sacrebleu\footnote{nrefs:1|case:mixed|eff:no|tok:13a|smooth:exp|version:2.0.0} \cite{papineni-etal-2002-bleu,post-2018-call} score. As the test set is not sentence aligned, we translate each document and then align the output sentences (hypothesis) to corresponding reference document using mwerSegmenter \citep{matusov-etal-2005-evaluating}, before calculating BLEU. 

For the baseline, we translate the test set using the segmentations produced by ASR.
 For our proposed window-based method, we evaluate using different fixed-size windows of length 
 8, 10, $\ldots$, 20 tokens. The results are shown in \Cref{tab:results:seg-vs-window}where  we observe that the proposed method outperforms the baseline with margins of 1.3 and 2.0 BLEU. 
These BLEU scores in \Cref{tab:results:seg-vs-window} across different window length are the best scores obtained after exploring different threshold ($r$) of match (refer to line 7 of Algorithm~\ref{algo:merge}). We show the BLEU scores for each threshold in Appendix \Cref{tab:results:window}.

\begin{figure}[ht]
     \centering
     \begin{subfigure}[b]{0.4\textwidth}
         \centering
         \includegraphics[width=\textwidth]{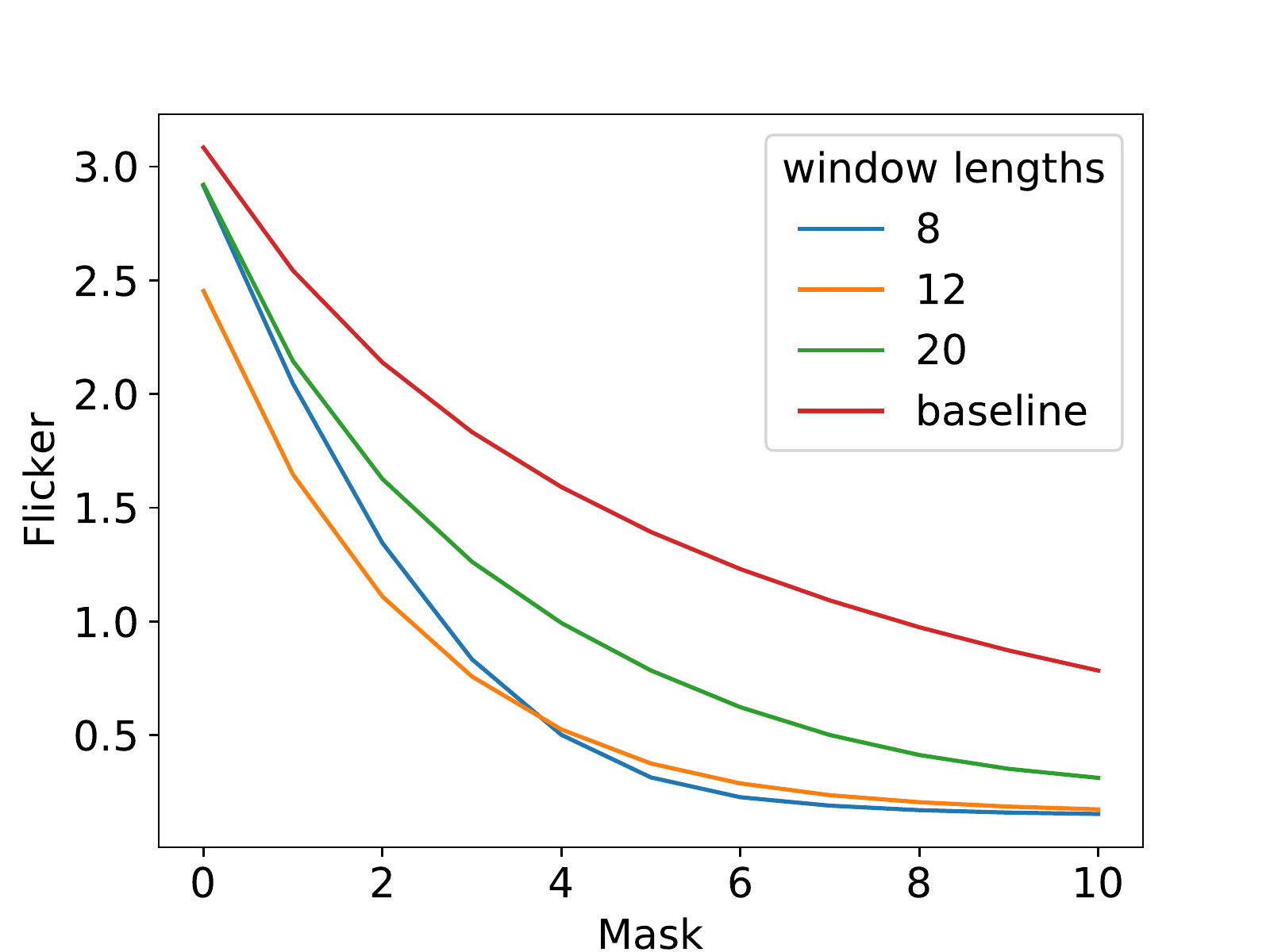}
         \caption{English-German}
         \label{fig:ende}
     \end{subfigure}
     \begin{subfigure}[b]{0.4\textwidth}
         \centering
         \includegraphics[width=\textwidth]{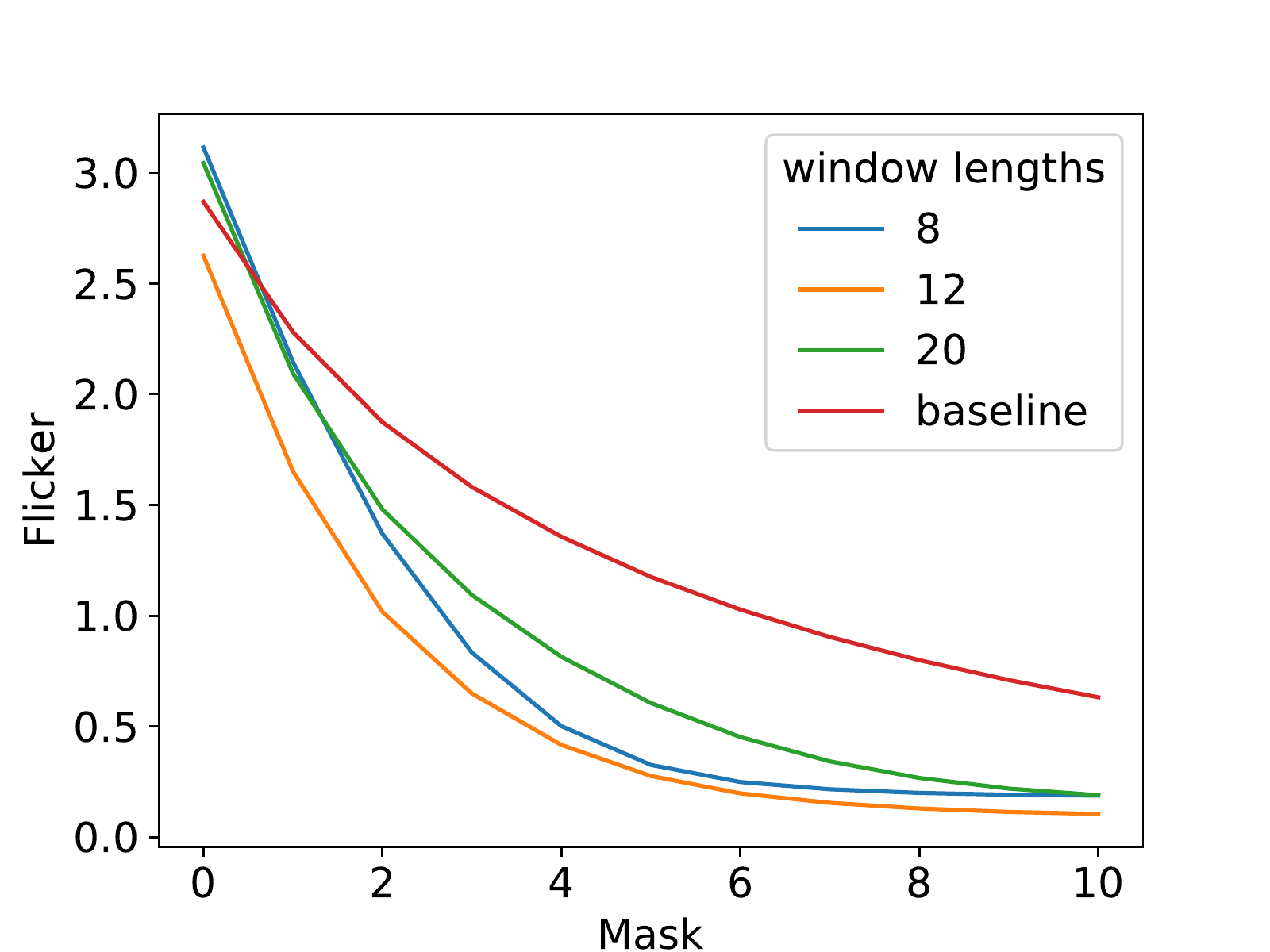}
         \caption{English-Czech}
         \label{fig:encs}
     \end{subfigure}
        \caption{Mask vs Flicker plots for different window lengths at threshold $r$ = 0.4, and the baseline.}
        \label{fig:plot:flicker-mask}
\end{figure}

For online SLT, since our system uses retranslation, we evaluate quality using BLEU, and flicker using normalised erasure (NE; \citealt{arivazhagan2019retranslation}). We first note that flicker is affected by both window length and thresholds -- shorter windows force commitment earlier which gives lower flicker. Low thresholds promote spurious matches, 
make translation flicker more, whilst high thresholds force too many retranslations, and will cause extra flicker when the maximum backoff is exceeded. After exploration (Appendix~\ref{app:match}), we set the threshold to 0.4 for the rest of our experiments.


\Cref{fig:plot:flicker-mask} shows the flicker-latency tradeoff of our sliding-window approach to 
online SLT, as we vary the fixed mask. We can see that the tradeoff is improved at all window sizes. This 
improvement is because the window approach only allows updates that are within the window length. The quality
of the online SLT (as measured on full sentences) is the same as the offline SLT. The flexible mask 
allows further improvements in flicker, for matched latency \BH{Refer to appropriate graph/table}

\section{Conclusion}
We proposed window-based approach which works at window (of fixed length of tokens) level, and removes the need of automatic sentence-segmentation of ASR output in cascaded SLT. We experimented with English-German and English-Czech language pairs and found that our proposed approach  
performs better than the segmentation based translation obtaining an improvement of 1.3-2 BLEU points. We also observed that masking the output reduced the flicker by a considerable margin as compared to the baseline.

    


\section*{Acknowledgements}
\lettrine[image=true, lines=2, findent=1ex, nindent=0ex, loversize=.15]{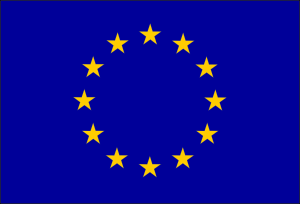} This work has received funding from the
European Union’s Horizon 2020 Research
and Innovation Programme under Grant Agreement
No 825460 (ELITR).

Ondřej Bojar would also like to acknowledge the support from
the 19-26934X (NEUREM3) grant of the Czech Science Foundation.

This work used the Cirrus UK National Tier-2 HPC Service at EPCC (http://www.cirrus.ac.uk) funded by the University of Edinburgh and EPSRC (EP/P020267/1)

\bibliography{anthology,biblio}

\clearpage
\appendix
\section{Training data statistics}
\label{app:training}

In \Cref{tab:data} we show the breakdown of our training data.

\begin{table}[ht]
\centering
\begin{tabular}{lr}
\hline
Corpus & Sentence pairs  \\
\hline
\multicolumn{2}{c}{English-German}\\
Europarl & 1.79 M  \\
Rapid & 1.45 M \\
News Commentary & 0.35 M \\
OpenSubtitle & 22.51 M \\
TED corpus & 206 K \\
MuST-C.v2 & 248 K \\
\hline
 \multicolumn{2}{c}{English-Czech} \\
 Europarl & 645 K\\
 ParaCrawl & 14 M\\
 CommonCrawl & 161 K\\
 News Commentary & 260 K\\
 CzEng2.0 & 36 M\footnote{deduped; original has 75 M}\\
 Wikititles & 410 K\\
 Rapid & 452 K\\
 
\hline
\end{tabular}
\caption{Corpora used in training the systems}
\label{tab:data}
\end{table}

\section{Creation of Windowed Parallel Corpus}
\label{app:window}

First, we word-align the pre-processed parallel corpus $D$ to obtain alignment $A$ using \texttt{fast\_align} \citep{dyer-etal-2013-simple}. Then we concatenate all the source-target sentence pairs $(s_k, t_k)$ into a single, very long, pair $(s, t)$ and subsequently, revise the alignment using the Algorithm~\ref{algo:revise}, so that the indexes are still correct in the 
concatenated corpus.

\begin{algorithm}[h]
\caption{Pseudo code for collapsing a word-aligned parallel corpus into a single pair of sentences preserving the word alignments.}
\label{algo:revise}
\begin{algorithmic}[1]
\REQUIRE Parallel corpus
  $D=\{(s_1, t_1),$ $(s_2, t_2),$ $\dots{},$ $(s_n, t_n)\}$,  alignment $A=\{a_1, a_2, ..., a_n\}, s=\epsilon, 
  t=\epsilon$,
  revised alignment $A^{'}=\{\}$
\FOR{$k \gets 1$ to $|D|$}

  \FOR {each $i, j \in a_k$}
    \STATE{$i \leftarrow i + |s|$} 
    \STATE{$j \leftarrow j + |t|$}
        \STATE{$A^{'} \leftarrow A^{'} \cup (i, j)$}
  \ENDFOR
  \STATE $s\leftarrow s + s_k$ \hfill\COMMENT{concatenation}\\
  \STATE $t\leftarrow t + t_k$ \hfill\COMMENT{concatenation}\\
\ENDFOR
\RETURN {$s, \quad t, \quad A^{'}$}
\end{algorithmic}
\end{algorithm}

Once we have combined the parallel corpus into a pair of sentences $(s,t)$, we use the revised alignment $A^{'}$ to generate parallel windows of length 15-25 tokens using Algorithm~\ref{alg:window}.

\begin{algorithm}[h]
\caption{Pseudo-code for extracting windows from the concatenated corpus}
\label{alg:window}
\begin{algorithmic}[1]
\REQUIRE Unsegmented source $s$, target $t$, and word alignment $A^{'}$ 
\STATE Initialize: $idx \leftarrow 0$
\WHILE{$idx < |t|$}
  \STATE {$l \leftarrow random(10, 25)$}
  \STATE $W_t \leftarrow t[idx: idx+l]$ \hfill\COMMENT{target window}
  \STATE $p = \min_i \{(i,j) \in A^{'}, idx \leq j < idx+l\}$
   \STATE $q = \max_i \{(i,j) \in A^{'}, idx \leq j < idx+l\}$ \\
  \STATE $W_s \leftarrow s[p:q]$ \hfill\COMMENT{source window}
  \STATE $idx \leftarrow idx + l$
\ENDWHILE
\end{algorithmic}
\end{algorithm}

\section{Exploration of Match Threshold}
\label{app:match}
\begin{figure}[ht]
 \centering 
 \includegraphics[scale=0.5]{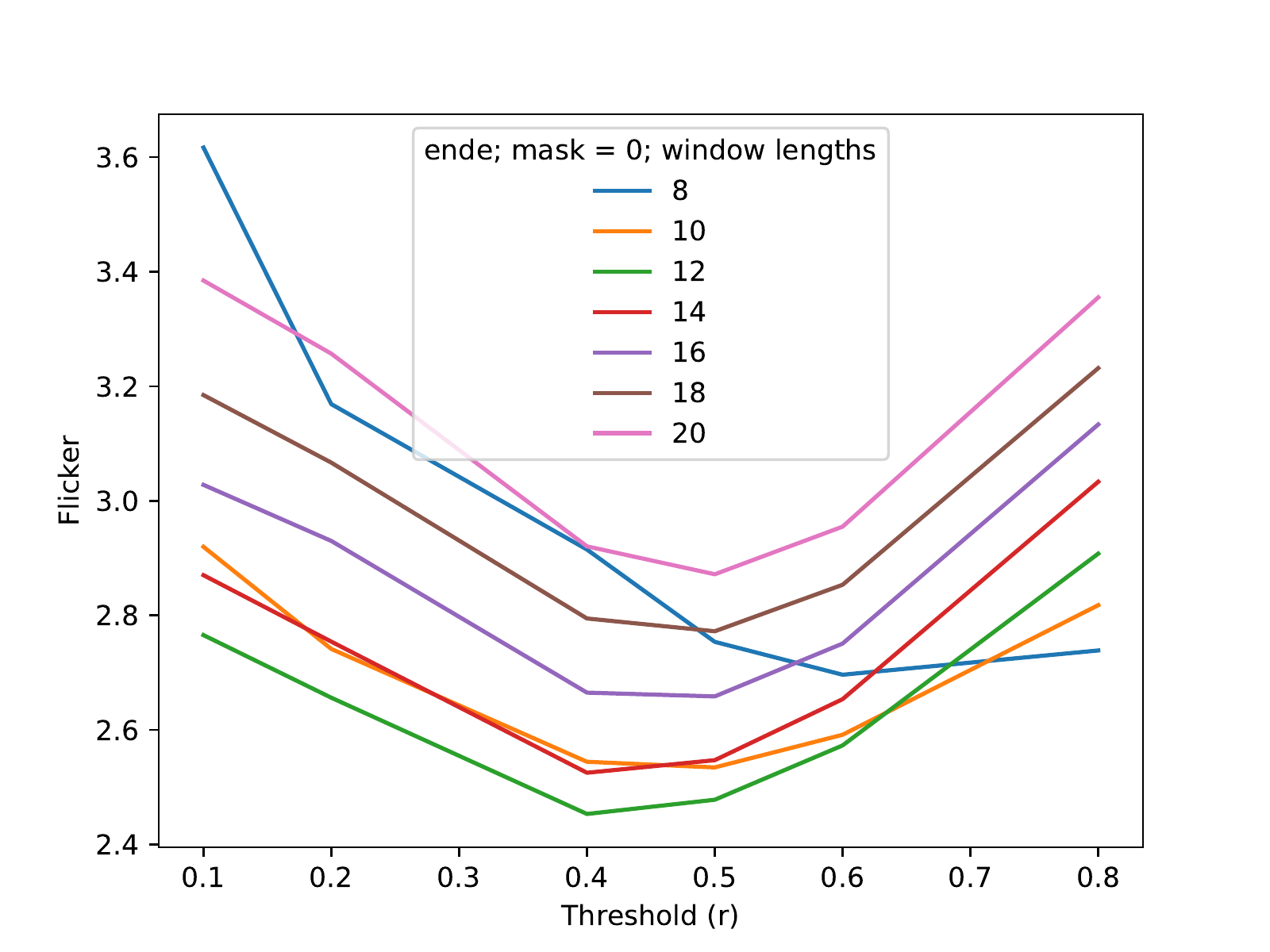}
 \includegraphics[scale=0.5]{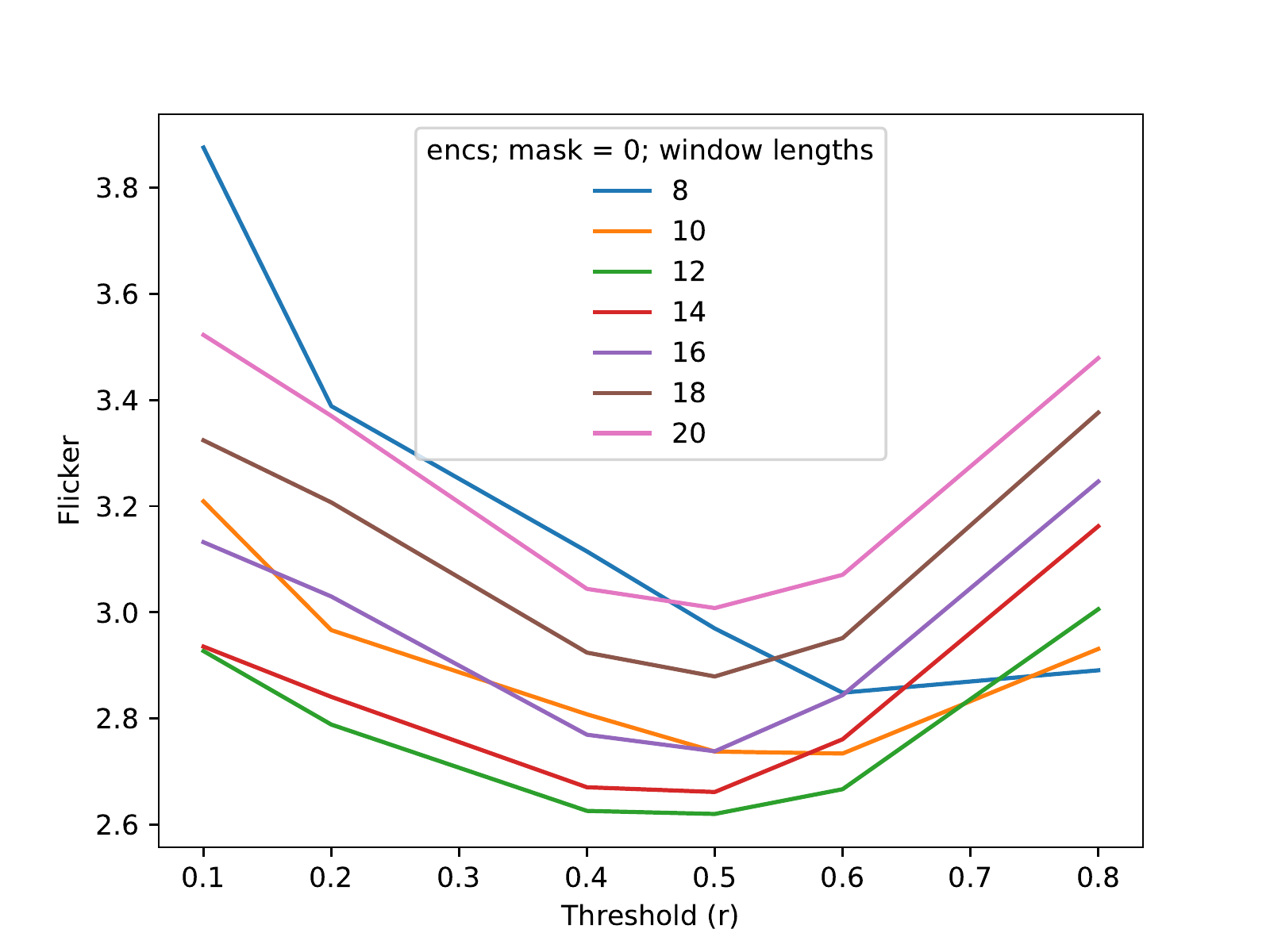}
 \caption{Threshold ($r$) vs Flicker plots. }
 \label{fig:thres_flicker}
\end{figure}

We have two hyperparameters to consider: \textit{window length} and \textit{threshold}, when generating the output.  We explore their combination to find the best threshold value. \Cref{tab:results:window} shows BLEU scores with different window length and threshold. We plot the flicker against the threshold for each window in \Cref{fig:thres_flicker} and we found 0.4 to be the best choice for threshold. Shorter windows force commitment earlier producing lower flicker. Low thresholds promote spurious matches making translation flicker more, whilst high thresholds force too many retranslations. We have shown the number of retranslation in \Cref{tab:results:mismatch} for different combination of window length and threshold. The reason why higher threshold forces too many retranslations is that even if we set higher threshold, it matches only with the match ratio between 0.5 to 0.6 on average. We have shown the average match ratio after joining every combination of window length and threshold in \Cref{tab:results:final_r}. We observe in \Cref{fig:thres_flicker} that higher threshold increases the flicker. The reason is that: as mentioned before, in one hand, it never reaches a match of $>$ 0.6 on average thus it retranslates more and generates longer output window, on the other hand, flicker depends on actual number of token mismatch - longer window will have more mismatch for the same threshold. In addition to that, these extra retranslations incurs an increase in computation requirement. However, this increase in complexity can be easily ignored, as in real life settings, largest source of latency is waiting for new source content from the speaker \citep{arivazhagan-etal-2020-translation}. 

\begin{table}[ht]
    \small
    \centering
    \begin{tabular}{l|cccccc}
        \hline
        & \multicolumn{5}{c}{Match Threshold ($r$)} \\
        Window($w_l$) & 0.1 & 0.2 & 0.4 & 0.5 & 0.6 & 0.8 \\
        \hline
        \multicolumn{7}{c}{en$\rightarrow$de} \\
		8 &  10.8 &  11.3 &  12.3 &  12.5 &  12.4 &  12.5 \\
		10 &  12.0 &  12.3 &  12.7 &  12.8 &  12.8 &  12.7 \\
		12 &  12.5 &  12.7 &  12.9 &  13.0 &  13.0 &  12.9 \\
		14 &  12.7 &  12.8 &  13.0 &  13.0 &  12.9 &  12.8 \\
		16 &  12.9 &  12.9 &  13.1 &  13.1 &  13.1 &  13.0 \\
		18 &  13.0 &  13.0 &  13.2 &  13.2 &  13.2 &  13.1 \\
		20 &  13.1 &  13.0 &  13.2 &  13.2 &  13.2 &  13.2 \\
        \multicolumn{7}{c}{en$\rightarrow$cs} \\
		8 &  8.3 &  9.1 &  9.8 &  10.0 &  10.0 &  9.9 \\
		10 &  9.5 &  9.7 &  10.2 &  10.3 &  10.2 &  10.2 \\
		12 &  10.0 &  10.2 &  10.4 &  10.4 &  10.4 &  10.4 \\
		14 &  10.2 &  10.4 &  10.5 &  10.4 &  10.5 &  10.4 \\
		16 &  10.5 &  10.6 &  10.6 &  10.6 &  10.6 &  10.5 \\
		18 &  10.5 &  10.5 &  10.6 &  10.6 &  10.6 &  10.5 \\
		20 &  10.5 &  10.6 &  10.7 &  10.7 &  10.5 &  10.5 \\
        \hline
    \end{tabular}
    \caption{Results with different window length and threshold. Sacrebleu computed after sentence aligning each document using mwerSegmenter. Bleu scores in green have the lowest flickers.}
    \label{tab:results:window}
\end{table}

\begin{table}[ht]
  \small
  \setlength{\tabcolsep}{3pt}
    \centering
    \begin{tabular}{l|ccccccl}
        \hline
        & \multicolumn{5}{c}{Match Threshold ($r$)} \\
        $w_l$ & 0.1 & 0.2 & 0.4 & 0.5 & 0.6 & 0.8 & \#windows\\
        \hline
        \multicolumn{7}{c}{en$\rightarrow$de} \\
		8 & 1724 & 10513 & 66471 & 103034 & 140889 & 200441 & 45879 \\
		10 & 1303 & 7352 & 50345 & 82775 & 118991 & 185398 & 45497 \\
		12 & 956 & 6394 & 46528 & 74698 & 110669 & 178805 & 45115 \\
		14 & 702 & 4809 & 42886 & 69847 & 105207 & 173017 & 44733 \\
		16 & 432 & 4098 & 40447 & 66391 & 100591 & 167585 & 44351 \\
		18 & 308 & 3809 & 38774 & 65132 & 99410 & 163935 & 43969 \\
		20 & 215 & 3407 & 37358 & 64266 & 96701 & 162025 & 43587 \\
        \multicolumn{7}{c}{en$\rightarrow$cs} \\
		8 & 2388 & 14757 & 74465 & 111900 & 148605 & 206238 & 45879 \\
		10 & 1257 & 8906 & 53651 & 84838 & 120135 & 188964 & 45497 \\
		12 & 1374 & 7170 & 44905 & 71432 & 105294 & 176580 & 45115 \\
		14 & 1094 & 5825 & 40480 & 64564 & 97436 & 169418 & 44733 \\
		16 & 806 & 4762 & 37067 & 60457 & 92346 & 163384 & 44351 \\
		18 & 489 & 4118 & 34710 & 58321 & 89392 & 158114 & 43969 \\
		20 & 292 & 3807 & 33440 & 57187 & 87418 & 154827 & 43587 \\
        \hline
    \end{tabular}
    \caption{Number of extra retranslations due to history extension. $w_l$ is window length. }
    \label{tab:results:mismatch}
\end{table}

\begin{table}[ht]
  \small
  \setlength{\tabcolsep}{3pt}
    \centering
    \begin{tabular}{l|ccccccl}
        \hline
        & \multicolumn{5}{c}{Match Threshold ($r$)} \\
        $w_l$ & 0.1 & 0.2 & 0.4 & 0.5 & 0.6 & 0.8 & \#windows\\
        \hline
        \multicolumn{6}{c}{en$\rightarrow$de} \\
        8&0.40&0.42&0.52&0.56&0.57&0.55&45879 \\
        10&0.47&0.49&0.57&0.59&0.60&0.58&45497 \\
        12&0.51&0.52&0.59&0.62&0.62&0.59&45115 \\
        14&0.52&0.54&0.60&0.63&0.64&0.60&44733 \\
        16&0.53&0.55&0.61&0.64&0.65&0.62&44351 \\
        18&0.54&0.55&0.61&0.64&0.65&0.62&43969 \\
        20&0.55&0.56&0.62&0.64&0.65&0.62&43587 \\
        \multicolumn{6}{c}{en$\rightarrow$cs} \\
        8&0.37&0.41&0.51&0.54&0.56&0.54&45879 \\
        10&0.45&0.47&0.56&0.59&0.60&0.57&45497 \\
        12&0.50&0.52&0.59&0.61&0.63&0.60&45115 \\
        14&0.53&0.54&0.61&0.63&0.65&0.61&44733 \\
        16&0.54&0.55&0.62&0.64&0.65&0.63&44351 \\
        18&0.55&0.56&0.62&0.65&0.66&0.63&43969 \\
        20&0.56&0.57&0.63&0.65&0.66&0.64&43587 \\
        \hline
    \end{tabular}
    \caption{Average match ratio after joining all the windows across different window length and threshold. We define average match ratio as $ \frac{1}{\#window}\sum^{\#window}\frac{\mathrm{match\_length}}{\mathrm{output\_window\_length}}$}
    \label{tab:results:final_r}
\end{table}




\end{document}